\documentclass{article}

\usepackage{PRIMEarxiv}

\usepackage[utf8]{inputenc} 
\usepackage[T1]{fontenc}    
\usepackage{hyperref}       
\usepackage{url}            
\usepackage{booktabs}       
\usepackage{amsfonts}       
\usepackage{nicefrac}       
\usepackage{microtype}      
\usepackage{lipsum}
\usepackage[numbers]{natbib}
\usepackage{bm}
\usepackage{fancyhdr}       
\usepackage{graphicx}       
\graphicspath{{media/}}     

\title{A Markovian Formalism for Active Querying
}

\author{
  Sid Ijju \\
  University of California Berkeley \\
  \texttt{sidijju@berkeley.edu} \\
}
\begin{document}
\maketitle

\begin{abstract}
Active learning algorithms have been an integral part of recent advances in artificial intelligence. However, the research in the field is widely varying and lacks an overall organizing leans. We outline a Markovian formalism for the field of active learning and survey the literature to demonstrate the organizing capability of our proposed formalism. Our formalism takes a partially observable Markovian system approach to the active learning process as a whole. We specifically outline how querying, dataset augmentation, reward updates, and other aspects of active learning can be viewed as a transition between meta-states in a Markovian system, and give direction into how other aspects of active learning can fit into our formalism.
\end{abstract}


\section{Introduction}
Current reinforcement learning policies are heavily dependent upon the usage of a reward function. However, for advanced problems such as learning behavior for a self-driving car or a robot with many degrees of freedom, the reward function is very difficult or impossible to design to effectively encompass all human considerations. Inverse reinforcement learning algorithms are a class of algorithms that attempt to solve this issue by learning a reward function from expert demonstrations, and then subsequently using any manner of standard reinforcement learning algorithms to learn an optimal policy based off that learned reward function \cite{ng2000}. 

Inverse reinforcement learning algorithms require a dataset of expert demonstrations to derive a reward function from. However, this reward function can often be inaccurate and highly variable. Furthermore, the learned reward function can often be misaligned with the true intentions of the human user, even if it does accurately account for the demonstrated data it learned from \cite{bobu2023aligning}. One recent advance in the training of inverse reinforcement learning algorithms is to use additional expert data about preferences between demonstrations \cite{settles2012}. The general idea is to first train an inverse reinforcement learning algorithm on a dataset of demonstrations, and then subsequently actively query a human expert to select a preference between generated demonstrations based off the learned prior of trajectories from the dataset of demonstrations. The policy can then be updated using the new expert data. 

We identify two general categories of research direction within the field of active learning. First, one could learn how to more effectively utilize the information provided by the human expert with each query. Second, one could learn how to increase the amount of information provided by each individual query. We can call the two directions information derivation (for learning how to maximize the information derived from a single query), and information optimization (for learning how to ask more optimal future queries) respectively. Information optimization approaches can be further divided into approaches that take a policy/learning algorithm-based approach as opposed to a data-based approach. 

The key observation of our work is that both research directions can be viewed as an \textit{informational choice} made by the robot at each step. More formally, we can view the active learning querying process itself as a Markov Decision Process (MDP) with non-Markovian reward. The action space and the transition function encompass both the choices of \textit{information type} the robot can choose to query as well as the \textit{information derived} by the robot from the query. In other words, both directions of research focus on \textbf{determining the best setup for the querying MDP}. 

We organize the field of query-based active learning using this formalism. Furthermore, we outline future research directions using the proposed formalism and discuss applications of the outlined methods. 

\section{Formalism}
\label{sec:formalism}

\paragraph{Model} We model our actual learning environment as a deterministic, fully-observable Markov decision process. We define a set $\mathcal{S}$ to represent the state of the system, and a set $\mathcal{A}$ to denote the action input to the system. At a time $t$, the system will be at a state $s_t \in \mathcal{S}$ and the robot will be tasked with taking an action $a_t \in \mathcal{A}$. We further define a trajectory $\xi \in \Xi$ as a finite sequence of states and actions (we define the horizon $\mathcal{H}$ of a trajectory $\xi$ as the maximum number of states). Therefore, we can write a trajectory $\xi = ((s_t, a_t))^T_{t=0}$, where $\max_{\xi \in \Xi}{T} = \mathcal{H}$. We are provided an initial dataset $\mathcal{D} \subset \Xi$ of trajectories $\xi$. Finally, we assume a reward function $r \in \mathcal{R}: \Xi \rightarrow \mathbb{R}$ that encapsulates how a human wants the system to behave. The goal of inverse reinforcement learning is to learn the reward function $r$ that best represents the human intentions shown in the dataset $\mathcal{D}$. \\

We also make a common simplifying assumption about the reward function, $r$ \cite{jorden2019}. We let $r$ be a linear combination of selected features $\Phi: \Xi \rightarrow \mathbb{R}^d$. Now, we can write $r(\xi) = \omega^T \Phi(\xi)$. With this simplification, we no longer need to learn a potentially high dimensional $\mathcal{R}$ and instead can learn a finite dimensional ($d$) vector $\omega$ that describes the human's reward function.

\paragraph{Markovian Formulation} The central aspect of the active learning process is the idea of querying the human expert. To define and formalize the notion of a query, we first define an "information Markov Decision Process". We can then represent the querying process as an aspect of this new system. 

We define the state space $\mathcal{S}' \in \Xi \times \mathcal{R}$. The state of our system at time will consist of a dataset $\mathcal{D}$ and a reward function $r \in \mathcal{R}$. We further define the action space $\mathcal{A'}$ as the set of potential queries (we use the term action and query interchangeably moving forward) we can ask using the current state $s \in \mathcal{S}'$ \footnote{For our discussion on the scope of the domain for $\mathcal{A}'$, refer to \ref{sec:actions}}, an observation space $\Omega$ that represents the set of human responses to a "query" $a \in \mathcal{A'}$, and a set of conditional observation probabilities $\mathcal{O}: \Omega \times S' \times A' \rightarrow [0, 1]$ that represent the probability of a response $o \in \Omega$ given the action $a \in \mathcal{A}'$ and the new state $s' \in \mathcal{S}'$. We also have $\mathcal{R}': \mathcal{S}' \times \mathcal{A}' \rightarrow \mathbb{R}$ and $\mathcal{T}': \mathcal{S}' \times \mathcal{A}' \rightarrow \mathcal{S}'$, and $\gamma$ (the discount factor). In this new system, the reward function $\mathcal{R}'$ represents the benefit of asking our query for the state of the system and its capability to learn the original reward function $r \in \mathcal{R}$. Furthermore, the transition function encapsulates the changes in system information that receiving a response to a query (action) entails.\\

Define our "information Markov Decision Process" as a partially observable Markov Decision Process $\mathcal{M}' = (\mathcal{S}', \mathcal{A}', \mathcal{T}', R', \Omega, \mathcal{O}, \gamma)$. Our claim is that \textbf{you can represent any active querying algorithm as a policy that optimizes on an "information Markov Decision Process" }.
\section{Survey}
\label{sec:survey}

We now survey the active learning literature using the formalism above with different selections of $\mathcal{A'}, \mathcal{T'}, \mathcal{R'}, \mathcal{O}$. Many methods fall into a single one of the above categories. However, there are some other approaches that are more subtle and which require multiple changes across the different categories defined to fit under the formalism, speaking to the nontrivial nature of identifying a common formalism for this field.

\subsection{Action Space: \texorpdfstring{$\mathcal{A'}$}{A'}}
\label{sec:actions}

To preface this section, we identify a simplifying assumption used by our formalism. We define the set $\Omega$ as the set of possible human responses to a "query" $a' \in \mathcal{A}'$ which uniquely ties the definition of the set $\Omega$ to the definition of $\mathcal{A}'$, and so in all further references we will discuss $\mathcal{A'}$ and $\Omega$ in a joint context. Furthermore, any modification to the action space $\mathcal{A}'$ will require a modification to the transition probability space $T'$ since a transition function $T(s' | s, a)$ inherently depends on the action $a \in \mathcal{A'}$, so we also jointly discuss $\mathcal{T'}$. For simplicity, we will consider the transition function $T$ as a complete description of the set $\mathcal{T}'$ in this section. Since the transition function $T$ operates as a tuple, we can decompose the transition function using a dataset transformation function $F \in \mathcal{F}: \Xi \rightarrow \Xi$ and a learning algorithm $L \in \mathcal{L}: \Xi \times \mathcal{R} \rightarrow \mathcal{R}$ that updates a learned reward function $r \in \mathcal{R}$ using a dataset $\mathcal{D} \in \Xi$. 

Most research in this field considers changes to the algorithm used for selection of the action $a \in \mathcal{A}'$. Barring trivial selection strategies such as random selection, uncertainty sampling remains the most popular strategy for this application \cite{lewis1994}, which involves simply selecting the sample the training algorithm is least certain about for labeling. Other strategies include query-by-committee \cite{seung1992} and expected model change \cite{settles2008}, among others \cite{settles2009active, phillips}. Some research also considers the ease of answering by the human for the query within the querying strategy \cite{easy}. Below, we will discuss the changes made to the system itself, but note that most if not all of the considered research above focuses heavily on the algorithm to solve the "information MDP" and not the changes required to the actual Markov system itself. For the full list of considered changes to the Markov model to account for changes in query type that will be discussed in this section, refer to \ref{tab:actiontable}.

\begin{table}
 \caption{The choice of $\mathcal{A'}$, and $T$ for different types of queries as described in \ref{sec:actions}}
  \centering
  \begin{tabular}{lll}
    \toprule
    \textbf{Query Type} & $\bm{\mathcal{A'}}$ & $\bm{T(s'|s, a)}$ \\
    Labels & $\xi \notin \mathcal{D}$ & $I(s' = (\mathcal{D}', L(\mathcal{D}', r)), \mathcal{D}' = \mathcal{D} \cup \{\xi\}$\\
    Comparisons & $(\xi_a, \xi_b) \in \mathcal{D}$ & $I(s' = (\mathcal{D}, L(D, r))$\\
    Demonstrations & $s \in S$ & $I(s' = (\mathcal{D}', L(\mathcal{D}', r)), \mathcal{D}' = \mathcal{D} \cup \{o\}$\\
    Feature Labels & $f(\xi)_i$ & $I(s' = (\mathcal{D}, L(D, r))$\\
    Language & $\xi \in \mathcal{D}$ & $I(s' = (\mathcal{D}', L(\mathcal{D}', r)), \mathcal{D}' = (\mathcal{D} \setminus \{\xi\}) \cup \{o\}$\\
    \bottomrule
  \end{tabular}
  \label{tab:actiontable}
\end{table}

\paragraph{Labels} The vast majority of traditional AL methods \cite{unlabeled1, unlabeled2, unlabeled3} utilize a simple querying strategy. The algorithm initially learns from a dataset $\mathcal{D} \subset \Xi$. Then, the algorithm selects an unlabeled instance $\xi \notin \mathcal{D}$ (or several \cite{pool, batch, sener2018active}) - this is our definition of the action space $\mathcal{A}'$. The algorithm also requests a label for the selected sample - this label can be a binary indication of the selected samples quality or in some cases  \cite{classification} a more fine-grained rating such as classifications or numerical scales. The choice of rating defines the set $\Omega$ - since there are many variations of labeling methods spanning discrete, continuous, categorical, and numerical sets, we refrain from exactly specifying the possible values for $\Omega$. We do note that the selection of $\Omega$ highly varies from application to application but in general simply encompasses a measure of quality for the samples $\xi \in \mathcal{D}$. After the algorithm receives the response, it will add the selected sample to its dataset $\mathcal{D}$ and relearn the reward function $r \in \mathcal{R}$, which can be stated as a deterministic transition between two states in our new system. \\

This avenue has received much attention in the active learning literature due to the simplicity of the action space - the complexity of strategies utilizing this type of action space usually falls within the query selection algorithm \cite{label1}. However, recently more methods have begun to propose changes to the action space itself \cite{cakmak12}, paving the way for more intelligent query selection algorithms by moving complexity into the action space. 

\paragraph{Comparisons} We define a comparison as a type of query where the human responder is given two trajectories $\xi_a, \xi_b$ and is asked to select the better trajectory of the two. Comparison queries have received wide study in literature due the removal of the need to maintain an online learning procedure \cite{comp1, comp3}. Formally, the dataset $\mathcal{D}$ remains constant throughout, and the function $L(\mathcal{D}, r)$ can be modified to only update the reward function $r$ as a result of the response to the given query. For example, one could update $r$ to minimize the probability of $\xi_b$ receiving a high reward while maximizing the probability of $\xi_a$ receiving a high reward (assuming $\xi_a$ was preferred to $\xi_b$). Under the linearity assumption, this becomes even simpler and can be succinctly written as a gradient update on the learned human reward vector $\omega$. \\

However, comparison query methods tend to suffer from lack of information value - the value of a binary comparison between two queries depends largely on the queries selected. Furthermore, due to the fixed nature of the dataset $\mathcal{D}$ used to select queries, the improvement of the active learning algorithm depends heavily on the original dataset $\mathcal{D}$. Other research attempts to address this by utilizing $K-$wise comparisons and other more informational comparisons such as rankings \cite{comp2}. Another interesting approach in this vein is one where the comparison is between two candidate labels instead of two candidate trajectories \cite{comp4}. We refrain from including all manner of specifications of the action space in Table \ref{tab:actiontable} due to the similarity they maintain to the default comparison specification but all can be fit in under our formalism.

\paragraph{Demonstrations} There is also a large body of work on the task of actively querying for demonstrations themselves \cite{dempref1, dempref2, demonstrations}. Instead of relying on a set of trajectories to ask for labels from, this type of querying asks the human to generate a trajectory from a starting state provided by the robot. In contrast to comparison queries, which use more features in the action space, this approach minimizes action space features and instead moves complexity to the response space $\Omega$, which in this case would be equivalent to the space of all trajectories $\Xi$. 

An interesting extension to this work is the idea of providing multiple states for the human to use of waypoints in the requested trajectory generation \cite{keyframe}. This method simply uses a larger action space (several states instead of one) but still follows the same reasoning for fitting under the formalism as above.

\paragraph{Feature Labels} Although feature labels as a query type are a relatively new addition to the research body in the field of active learning, they have still been the subject of several studies \cite{cakmak12, feature1, feature2}. The general approach involves querying the human expert about the usefulness of a particular feature in relation to a data point or overall objective. This methodology takes advantage of the fact that in practice, many active learning algorithms opt to use a feature transformation $f: \Xi \rightarrow \mathbb{R}^d$ where $d << \mathcal{H} * dim(\mathcal{S}) * dim(\mathcal{A})$. We query about the usefulness of one of the dimensions $d$ and one of its corresponding values towards the overall objective and use the human response to reweight our model's learned importance of the feature $d$. This can essentially be viewed as a transformation on the learning algorithm $L$ to consider a weighted version of the dataset $\mathcal{D}$. 

\paragraph{Language} Many recent advances in active learning have been in the domain of natural language processing \cite{lang2, lang3}. Specifically, Reinforcement Learning from Human Feedback algorithms for language models has been tremendously successful. In this body of work, the requested query for the human labeler is either a selection between different language outputs or a correction on a generated output \cite{lang1}. If we view sentences as a type of trajectory, we can view these options for actions as either a comparison (which we have already covered) or a correction. We can formalize a correction for a trajectory query $\xi$ as essentially the creation of a new trajectory $\xi'$. The dataset $\mathcal{D}$ will then be modified to replace $\xi$ with $\xi'$. Modifying an existing element of the dataset as part of the transition function lends this category of action a unique place under our framework. 
\subsection{Reward Space: \texorpdfstring{$\mathcal{R'}$}{R'}}
\label{sec:rewards}

Implicitly, many methods do contain specifications of the reward space $\mathcal{R}'$. Methods like expected model change and variance reduction, while primarily created for the purpose of action selection, could also feasibly be used as a measure of the reward a given action has on the state of the system. In contrast, there are relatively few methodologies that actively specify the reward space $\mathcal{R}'$. One method is information gain, used in \cite{dempref1, dempref2} which essentially computes the expected gain in information of a query given a distribution over the predicted response and the actual response. 

\subsection{Observation Probability Space: \texorpdfstring{$\mathcal{O}$}{O}}
\label{sec:obsprobs}

Most of the literature on active learning utilizes a simple model for human rationality that has received widespread acceptance in the field - the Boltzmann rationality model \cite{Boltzmann}. This model assumes that the human's choice of response will be proportional to the exponentiated return for that response. Under our formalism, any modification to the modeling used for modeling the probability of a human response would fall under the domain of $\mathcal{O}$. As such, this portion of the Markov system encompasses a large variety of research across fields such as psychology and computer science. 

\section{Discussion}
\label{sec:dis}

For all aspects of active learning, instead of viewing each aspect of the overall system as separate, our formalism ties them all together under the partially observable Markovian framework. the robot can extract information from humans by modeling them as making approximate reward-rational choices. We believe this unifying lens can enable us to better understand how to contend with all these system components. In the future, we hope this formalism will enable research on active querying with more integrated and complex learning algorithm as well as making it easier to do active learning research in new directions. 

In our survey, we discuss the body of work in the active learning field. We observe that the vast majority of work in the field occurs in the context of the action space $\mathcal{A}'$. In most cases, the transition function $\mathcal{T}'$ and $\Omega$ follow naturally as a result of the choice of $\mathcal{A}'$. We believe that many promising directions of research in the future could occur by accounting for the other aspects of the Markovian formalism, i.e. $\mathcal{R}'$ and $O$. Below, we outline some potential considerations and future directions of work.

\paragraph{Non-deterministic \texorpdfstring{$\mathcal{T'}$}{T'}} In our discussion of query types in \ref{sec:actions}, we can observe that the transition function is deterministic in all of the examples covered. However, it does not necessarily have to be. Under our formalism, we can propose several new avenues of research in terms of creating non-deterministic transitions between states in $\mathcal{S}'$. There are two main avenues to approach this. \\ \\
First, one could develop methods that add data to $\mathcal{D}$ in a non-deterministic manner e.g. use a parameter $\alpha$ as a probability threshold for adding a new query and response pair to the dataset $\mathcal{D}'$. If we treat this parameter as part of the query selection algorithm, we could effectively treat $\alpha$ as a selection metric for how valuable a query could be to ask a human expert. Normal methods also require many human samples to function properly - this approach allows for consideration for the human expert's time by probabilistically determining whether or not a given query would be valuable to the learning process as a whole. \\ \\
Second, we could also conceive of methods that non-deterministically change the learned reward function $r$ as a result of the changes to the state. If our learning algorithm $L$ has an element of randomness such that the output of the update $L(\mathcal{D}, r)$ is not deterministic, then we could instead represent $r$ as a distribution over possible reward functions that $L$ could output as a result of a query and data update instead of just one possibility. 

\paragraph{Dataset Tuning} Consider the modifications to the dataset $\mathcal{D}$ that occur in the transition function for our examples above. In all of the aforementioned research, the changes either consisted of an addition or no change at all. A new research direction could involve modifying old examples using new data gained from querying - for example, reducing the labeled reward for a query deemed bad by the human expert. 

\paragraph{Meta-RL} Many methods ignore the specification of the reward space $\mathcal{R}'$ in this Markovian setup. Implicitly, most methods do however use some sort of quantitative assessment during the query selection phase. One potential future research direction could involve the design of a meta-RL algorithm that operates and optimizes on the "information MDP" by utilizing an approximation of query value as the reward function and then optimizing based on that. 

\section{Conclusion}

As the capability of artificial intelligence advances, it has become increasingly more important to develop a nuanced understanding of how intelligence systems learn and work. For all types of learning systems, from self-driving cars to assistive robots, we believe that robots cannot rely on simply specified reward functions and systems to guarantee long term user satisfaction. By considering and formalizing the full complexities of the active learning process, we believe we take a small step towards better understanding how learning processes can function as a whole, and we hope our work contributes to a future where in which robots fully take advantage of all the complexities inherent within their learning processes. We are optimistic that future research will account for the nuances in the formalism and seek out algorithms and modifications that fully take advantage of the complicated relationships within the Markovian system. Overall, we hope that our work offers a unifying perspective on the field of active learning as a whole and provides insight towards creating a truly intelligent robot.

\bibliographystyle{unsrt} 
\bibliography{references}

\end{document}